# Recursive Least Squares based Refinement Network for the Rollout Trajectory Prediction Methods

Qifan Xue, Xuanpeng Li*, Weigong Zhang

*Abstract*—Trajectory prediction plays a pivotal role in the field of intelligent vehicles. It currently suffers from several challenges, e.g., accumulative error in rollout process and weak adaptability in various scenarios. This paper proposes a parametric-learning recursive least squares (RLS) estimation based on deep neural network for trajectory prediction. We design a flexible plug-in module which can be readily implanted into rollout approaches. Goal points are proposed to capture the long-term prediction stability from the global perspective. We carried experiments out on the NGSIM dataset. The promising results indicate that our method could improve rollout trajectory prediction methods effectively.

## I. Introduction

Trajectory prediction is an important function in autonomous systems. History trajectories of neighboring agents are utilized to predict their future position, which helps ego agent with path planning. The challenge is that the trajectory tends to be highly non-linear due to uncertain behaviors and complex interactions of agents.

A number of studies have been conducted to model and predict the trajectory in traffic scenes. Recent work suggests to employ RNN (Recurrent Neural Network)[1] to encode the history trajectory. Besides, the social mechanism and the attention mechanism[2] are used to take agents' interactions into account. Moreover, VAE (Variational Auto Encoder)[3] and GAN (Generative Adversarial Networks)[4] are proposed to improve model's performance. Together, these studies take the rollout framework to forecast results iteratively. The term 'rollout' is used here to refer to the cyclical way to obtain predictions.

Nonetheless, most research up to now has not treated the two limitations in detail:

**The rollout collapse**: The rollout algorithms optimally update current states (position, velocity, etc.) at each time step, which loops to get the entire trajectory. However, the error accumulates due to the continuously updated states have a more significant impact on the next state. This problem is broadly defined as 'the rollout collapse'.

**The local optima problem**: The rollout algorithms always fail to make a long-term prediction because the algorithm is designed to run step by step, which makes the local optimal decision based on the past movement history and other agents' behaviors.

This paper proposes a lightweight plug-in module called 'Recursive Least Squares based Refinement Network'(RLSRN) to refine rollout models by integrating recursive least squares [19] in a plausible deep learning way. Recursive Least Squares (RLS) uses a series of measurements observed over time, containing statistical noise, and produces estimates of unobserved variables more precise.

As shown in Figure 1, goal points are defined as estimated target positions at several time steps. They are predicted in order to improve long-term prediction performance. At each time step, the probabilistic refinement module adjusts raw rollout

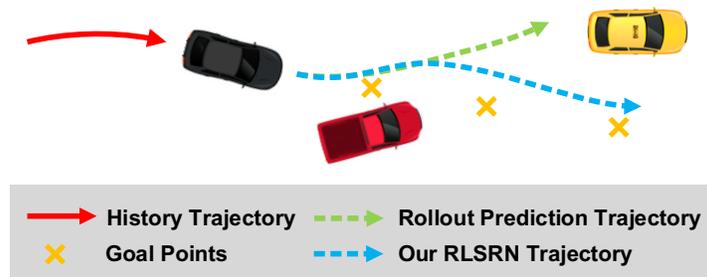

Figure 1. The introduction of RLSRN. The red line represents the past movement trajectory of target agent. The dotted green and blue lines respectively represent paths predicted by the rollout method and our method. The yellow crosses represent goal points which help to fix the rollout trajectory from the global perspective.

*Research supported by National Natural Science Foundation of China under Grant 61906038.

Qifan Xue (1995.6), Xuanpeng Li (1985.2) and Weigong Zhang (1959.10) are with the School of Instrument Science and Engineering, Southeast University, Nanjing, 210096, China (corresponding author: Xuanpeng Li, e-mail: li_xuanpeng@seu.edu.cn).

prediction with goal points according to the RLS principle. Experiments verify the performance of our method on the NGSIM dataset.

## II. RELATED WORK

A large and growing body of literatures have been investigated in the field of trajectory prediction since the development of the deep learning. RNNs and their variants, e.g., LSTM(Long Short-Term Memory)[6] and GRU(Gated Recurrent Neural Networks)[7] are widely used in the sequence estimation problem in lots of tasks such as localization[8] and decision making[9].

Besides, various models like VAE and GAN have been applied into multi-modal prediction. For example, Lee et al.[13] propose a conditional variational autoencoder (CVAE), named DESIRE, to generate multiple future trajectories based on agent interactions, scene semantics and expected reward function. Tang et al. [14] introduce a probabilistic framework named Multiple Future Prediction that efficiently learns latent variables to jointly model the multiple future motions of agents.

Additionally, numerous studies have been attempted to model social interactions. Social LSTM[10] introduces the social pooling mechanism to aggregate the neighboring agents' behaviors together. CS-LSTM[11] employs the convolution operation to improve the performance of the social modeling. Social GAN[12] integrates Social LSTM with the generative model.

In order to reduce the accumulative errors of rollout approaches, some previous research focus on setting the 'goal' or 'endpoint'. TPNet[15] makes final prediction by refining the trajectories which are generated by regressive endpoints and polynomial fitting. PECNet[16] infers distant trajectory endpoints to assist in the long-range multi-modal trajectory prediction. Goal-GAN[17] presents an interpretable end-to-end model for trajectory prediction with a goal estimation module and a routing module.

Previous research work mainly focuses on the model structure of deep neural networks, but lacks interpretability of physics mechanism. RLS is a recursive method of aggregating two or more random variables to minimize the variance of the sum[18], which could be used in aggregation of purposeful endpoints and past trajectories. To date, few studies have investigated the integration of RLS into the deep learning framework. Thus, this paper proposes a fusion framework based on the RLS and deep neural network to improve the interpretability of trajectory prediction based on deep learning methods.

## III. MODEL

### A. Problem Formulation

In this work, we tackle vehicle's trajectory prediction by formulating a probabilistic framework with N interacting agents in the traffic scene. Given the past trajectory of N agents as $X = \{X^1, X^2, \cdots, X^n\}$, where $X^i = \{x^i_{t-\tau}, x^i_{t-\tau+1}, \cdots, x^i_t\}$ denotes the n-th agent's position from time $t - \tau$ to t. The problem is how to predict the future positions $Y = \{Y^1, Y^2, \cdots, Y^n\}$, where $Y^i = \{y^i_t, y^i_{t+1}, \cdots, y^i_{t+T}\}$ denotes the position of all agents from time t to $t + T$ with T time steps.

The rollout methods factorize p(Y|X) as:

$$p(Y|X) = \sum_H p(Y, H|X) = \sum_H p(Y|H, X)p(H|X)$$

$$= \prod_{k=t+1}^{t+T} p(Y_k|Y_{t:k-1}, H_k, X) \, p(H_k|H_{k-1}, X)$$

where H denotes hidden variables representing temporal-spatial features.

However, there exists error accumulation problem along with accuracy decreasing over time. Hence, we introduce a set of stochastic variables $\Theta = \{\Theta^1, \Theta^2, \cdots, \Theta^n\}$, where $\Theta^i = \{\theta^i_t, \theta^i_{t+1}, \cdots, \theta^i_{t+T}\}$, to factorize the distribution as:

$$p(Y|X) = \sum_\Theta \sum_H p(Y, H, \Theta|X)$$

$$= \sum_\Theta \sum_H p(Y|H, \Theta, X)p(H|X)p(\Theta|X)$$

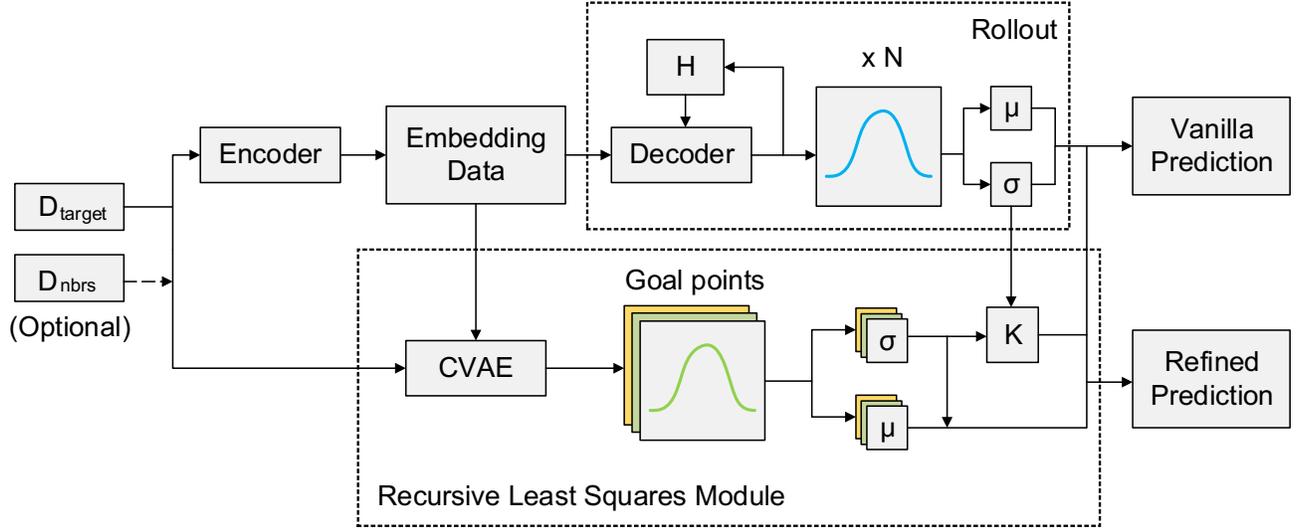

Figure 2. The pipeline of RLSRN. $D_{target}$ and $D_{nbrs}$ represent the trajectory of the target vehicle and neighboring vehicles respectively. H denotes hidden variables representing temporal features. The decoder of the rollout method loops N times for future prediction. RLSRN is devised as a plug-in component affiliated to the rollout method. Goal points are only predicted once at the beginning. In each step, the rollout trajectory is refined by RLS to generate the final prediction.

where $p(\Theta|X)$ can be further factorized as:

$$p(\Theta|X) = \prod_{k=t+1}^{t+T} p(\Theta_k|X)$$

The intuition is that Θ would represent driving intention to guide the rollout approach. The key of RLSRN is that Θ is only sampled at the time t, while it can provide information from the global perspective and overcome the accumulative error.

*B. RLS and Goal Points*

The vanilla Recursive Least Squares method updates current state variables, using weighted average with the measurement value interfered by noise, which can be formularized as:

$$K_k = P_{k-1} H_k^T (H_k P_{k-1} H_k^T + R_k)^{-1}$$

$$\hat{X}_k = \hat{X}_{k-1} + K_k (Z_k - H_k \hat{X}_{k-1})$$

$$P_k = (I - K_k H_k) P_{k-1}$$

where K denotes the gain. $\hat{X}$ represent the state estimation. P and R represent the covariance matrix of the estimation error and the measurement noise respectively. H is set to the identity matrix here. Z represents the measurement value.

As shown in Figure 2, $\hat{X}$ and P are generated by the rollout method. We introduce the estimated goal points to substitute for the measurement related components Z and R as following:

$$(\hat{X}, P), (Z, R) \sim \mathcal{N}(x, y, \sigma_x, \sigma_y, \rho)$$

$$(\hat{X}, P) = \phi(D; W_{rollout})$$

$$(Z, R) = \psi(D; W_{CVAE})$$

where $\phi(\cdot)$ represents the backbone rollout method. $\psi(\cdot)$ is the neural network implemented by CVAE. D is the trajectory data. W denotes weights learned in the neural network. After that, the estimation $(\hat{X}, P)$ can be calculated according to RLS principles.

*C. Recursive Least Squares Module*

We implement RLSRN in the form of a plug-in module with high compatibility. As shown in Figure 2, the encoder encodes the trajectories of target and neighboring agents for embedding. In the upper branch, the decoder of the rollout method outputs the mean and the variance of estimation. In the bottom branch, the RLS module produces goal points (both the value and the variance) once for all time steps. At each time step, the RLS module sequentially calculates the gain and the updated trajectory. Then, the rollout part takes the history trajectory with the latest prediction.

## IV. EXPERIMENTS

In this section, our method is evaluated on the public dataset NGSIM. NGSIM is a top-down camera recorded naturalistic trajectory dataset on US highway 101 (US-101) and interstate 80 (I-80). The dataset is split into 70% training set, 10% validation set, and 20% testing set. For each extracted 8 seconds trajectory segment, we use the last 3-second trajectory to predict the next 5-second trajectory. Raw data are sampled at the sampling rate of 10Hz. As the previous work[11][11][14], we downsample each segment by a factor of 2.

Throughout the training and testing process, the root mean squared error (RMSE) in meter, was used as the performance metric, which is defined as:

$$\text{RMSE} = \sqrt{\frac{\sum_{n=0}^{N}\sum_{t=0}^{T}((x_t^n - \hat{x}_t^n)^2 + (y_t^n - \hat{y}_t^n)^2)}{N \times T}}$$

We carried out the ablation experiments on the following methods:

RNN: vanilla LSTM model with the encoder-decoder framework.

RNN + Atten: Adding the dynamic attention mechanism to learn social interactions.

MFP[14]: the state-of-the-art model learning semantically meaningful latent variables.

### A. Qualitative Results and Analyses

Figure 3 illustrates the comparison of MFP and MFP + RLSRN in the same scene. The refined trajectory reveals that goal points act as the lighthouse on most occasions. The vanilla model sometimes falls into the wrong path but cannot be amended due to the rollout mechanism. Meanwhile, the refined model takes the initially generated 'goals' as a reference during the rollout process. In many cases, the trajectory based on the MFP+RLSRN yaws as well at the first several timesteps. Even though, RLSRN corrects the trajectory to a certain extent over time. Particularly, in Figure 3 (d), there exists an obvious wheeling in the trajectory based on the RLSRN method. Nonetheless, there is still room for further progress in avoiding mode-averaging to make the trajectory plausible.

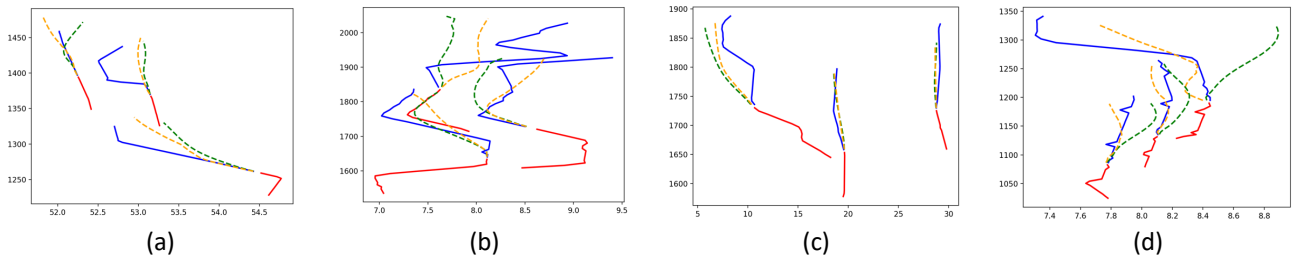

(a) (b) (c) (d)

Figure 3. Trajectory visualization. We present the prediction results of MFP and MFP + RLSRN in the same scene. The red line denotes the history path. The ground truth is plotted in blue line. The prediction trajectories of MFP and MFP + RLSRN are drawn in green and orange dotted lines. The axes of x and y are in meter.

### B. Quantitative Results and Analyses

Table 1 presents RMSE results of ablation experiments on three vanilla models: RNN, RNN+Atten and MFP with those refined by our RLSRN. RLSRN could reduce the RMSE at the many timesteps. In particular, there is a clear trend of refinement at the 5s with the improvement of 0.1m, 0.12m and 0.09m. What stands out is that RLSRN performs similarly on MFP compared with other models even if MFP has a state-of-art performance. Whereas no significant differences at the first 4s are found between the results of RNN and RNN+Atten, the RMSE at 5s decreases in both two models. Moreover, it can be seen in Figure 4 that RLSRN accelerates the convergence of model with the better performance in the whole process. Goal points may benefit the long-term prediction because initial intention helps to refine the rollout trajectory disturbed by accumulative errors.

## V. CONCLUSIONS

This paper sets out to deal with the rollout collapse and the local optima problem in rollout methods of trajectory prediction on road scenes. We extend RLS in a plausible deep learning way. The lightweight refinement module is implemented with high compatibility. Experimental results indicate an improvement of performance compared to the existing rollout methods.

This study is limited due to lack of the state formulation which extends the RLS estimation to the Kalman filter[19]. Further work needs to establish a tight coupling implement rather than the plug-in refinement module. Notwithstanding these limitations, the findings reported here shed new light on interpretability about deep learning methods.

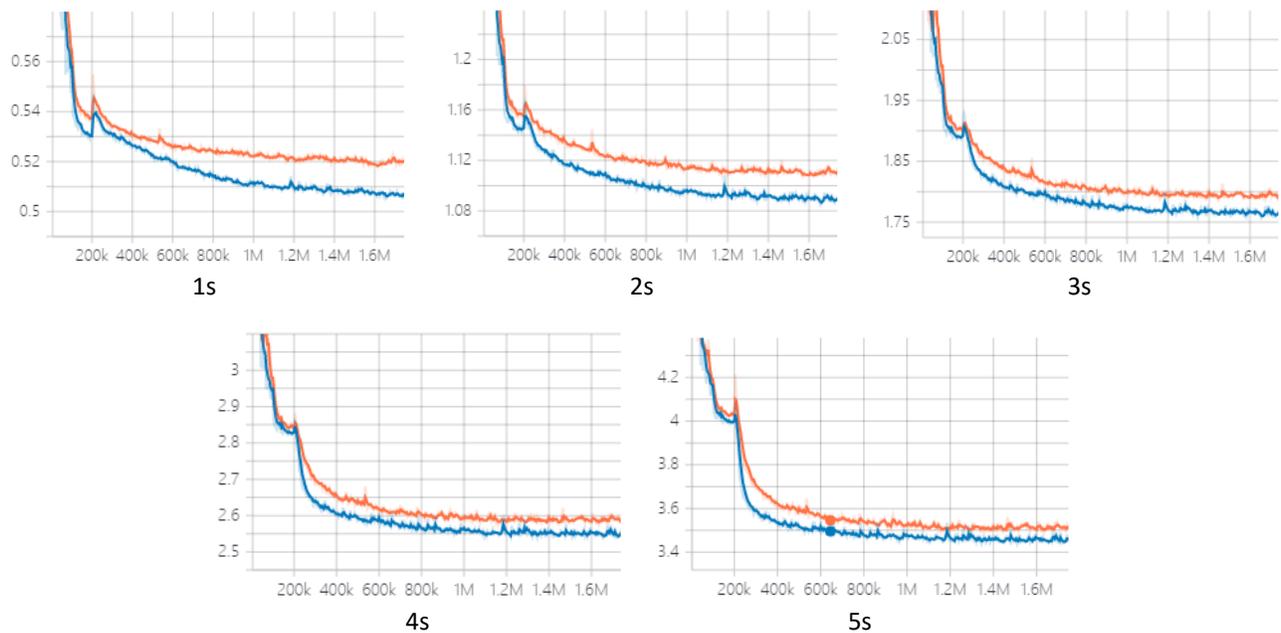

Figure 4. Evaluation curves of MFP and MFP + RLSRN module on test set. We pre-trained the model without interaction for the first 200k updates. The performance of MFP is shown in the orange line and that of MFP + RLSRN is shown in the blue line.